\pdfoutput=1

\documentclass[11pt]{article}

\usepackage{naacl2021}

\usepackage{times}
\usepackage{latexsym}
\usepackage{graphicx}
 \usepackage{color,soul}
\usepackage[T1]{fontenc}

\usepackage[utf8]{inputenc}

\usepackage{microtype}

\title{Production vs Perception: \\The Role of Individuality in Usage-Based Grammar Induction}

\author{Jonathan Dunn \\
  Department of Linguistics \\
  University of Canterbury \\
  \texttt{\small{\href{mailto:jonathan.dunn@canterbury.ac.nz}{jonathan.dunn@canterbury.ac.nz}}} \\\And
  Andrea Nini \\
  Linguistics and English Language \\
  University of Manchester \\
  \texttt{\small{\href{andrea.nini@manchester.ac.uk}{andrea.nini@manchester.ac.uk}}} \\}

\begin{document}
\maketitle
\begin{abstract}
This paper asks whether a distinction between production-based and perception-based grammar induction influences either (i) the growth curve of grammars and lexicons or (ii) the similarity between representations learned from independent sub-sets of a corpus. A \textit{production-based} model is trained on the usage of a single individual, thus simulating the grammatical knowledge of a single speaker. A \textit{perception-based} model is trained on an aggregation of many individuals, thus simulating grammatical generalizations learned from exposure to many different speakers. To ensure robustness, the experiments are replicated across two registers of written English, with four additional registers reserved as a control. A set of three computational experiments shows that production-based grammars are significantly different from perception-based grammars across all conditions, with a steeper growth curve that can be explained by substantial inter-individual grammatical differences.
\end{abstract}

\section{The Role of Individuals in Usage-Based Grammar Induction}

This paper experiments with the interaction between the amount of exposure (the size of a training corpus) and the number of representations learned (the size of the grammar and lexicon) under perception-based vs production-based grammar induction. The basic idea behind these experiments is to test the degree to which computational construction grammar \cite{Alishahi2008, Wible2010, Forsberg2014, d17, Barak2017, Barak2017a} satisfies the expectations of the usage-based paradigm \cite{g06a, Goldberg2011, Goldberg2016}. The input for language learning, \textit{exposure}, is essential from a usage-based perspective. Does usage-based grammar induction maintain a distinction between different types of exposure?

A first preliminary question is whether the grammar grows at the same rate as the lexicon when exposed to increasing amounts of data. While the growth curve of the lexicon is well-documented \cite{Zipf1935, Heaps1978, Gelbukh2001,Baayen2001}, less is known about changes in construction grammars when exposed to increasing amounts of training data. Construction Grammar argues that both words and constructions are \textit{symbols}. However, because these two types of representations operate at different levels of complexity, it is possible that they grow at different rates. We thus experiment with the growth of a computational construction grammar \cite{d18, Dunn2019} across data drawn from six different registers: news articles, Wikipedia articles, web pages, tweets, academic papers, and published books. These experiments are needed to establish a baseline relationship between the grammar and the lexicon for the experiments to follow.

The second question is whether a difference between perception and production influences the growth curves of the grammar and the lexicon. Most corpora used for experiments in grammar induction are aggregations of many unknown individuals. From the perspective of language learning or acquisition, these corpora represent a \textit{perception-based} approach: the model is exposed to snippets of language use from many different sources in the same way that an individual is exposed to many different speakers. Language perception is the process of hearing, reading, and seeing language use (being exposed to someone else's production). These models simulate perception-based grammar induction in the sense that the input is a selection of many different individuals, each with their own grammar.

This is contrasted with a \textit{production-based} approach in which each training corpus represents a single individual: the model is exposed only to the language production observed from that one individual. Language production is the process of speaking, writing, and signing (creating new language use). From the perspective of language acquisition, a purely production-based situation does not exist: an individual needs to learn a grammar before that grammar is able to produce any output. But, within the current context of grammar induction, the question is whether a corpus from just a single individual produces a different type of grammar than a corpus from many different individuals. This is important because most computational models of language learning operate on a corpus drawn from many unknown individuals (perception-based, in these terms) without evaluating whether this distinction influences the grammar learning process.

We conduct experiments across two registers that simulate either production-based grammar induction (one single individual) or perception-based grammar induction (many different individuals). The question is whether the mode of observation influences the resulting grammar's growth curve. These conditions are paired across two registers and contrasted with the background registers in order to avoid interpreting other sources of variation to be a result of these different exposure conditions.

The third question is whether individuality is an important factor to take into account in induction. On the one hand, perception-based models will be exposed to language use by many different individuals, potentially causing individual models to \textit{converge} onto a shared grammar. On the other hand, production-based models will be exposed to the language use of only one individual, potentially causing individual models to \textit{diverge} in a manner that highlights individual differences. We test this by learning grammars from 20 distinct corpora for each condition for each register. We then compute the pairwise similarities between representations, creating a population of perception-based vs production-based models. Do the models exposed to individuals differ from models exposed to aggregations of individuals?

The primary contribution of this paper is to establish the influence that individual production has on usage-based grammar induction. The role of individual-specific usage is of special importance to construction grammar: How much does a person's grammar actually depend on observed usage? The computational experiments in this paper establish that production-based models show more individual differences than comparable perception-based models. This is indicated by both (i) a significantly increased growth curve and (ii) greater pairwise distances between learned grammars.

\section{Methods: Computational CxG}

The grammar induction experiments in this paper draw on computational construction grammar \cite{d17, d18b, d18}. In the Construction Grammar paradigm, a grammar is modelled as an inventory of symbols of varying complexity: from parts of words (morphemes) to lexical items (words) up to abstract patterns (\textsc{np -> det n}). Construction Grammar thus rejects the notion that the lexicon and grammatical rules are two separate entities, instead suggesting that both are similar symbols with different levels of abstraction. In the same way as other symbols, the units of grammar in this paradigm consist of a \textit{form} combined with a \textit{meaning}. This is most evident in the case of lexical items, but also applies to grammatical constructions. For example, the abstract structure \textsc{np vp np np}, with the right constraints, conveys a meaning of transfer (e.g. \textit{Kim gave Alex the book}).

In order to extract a grammar of this kind computationally, an algorithm must focus on the form of the constructions. For example, computational construction grammars are different from other types of grammar because they allow lexical and semantic representations in addition to syntactic representations. On the one hand, this leads to constructions capturing item-specific slot-constraints that are an important part of usage-based grammar. On the other hand, this means that the hypothesis space of potential grammars is much larger. Representing the \textit{meaning} of these constructional forms is a separate problem from finding the forms themselves.

~

\hspace{1mm}(a) \textsc{np}-Simple -> \textsc{det} \textsc{adj} \textsc{n}

\hspace{1mm}(b) \textsc{np}-Construction -> \textsc{det} \textsc{adj} \textsc{[sem=335]}

\hspace{1mm}(c) ``the developing countries"

\hspace{1mm}(d) ``a vertical organization"

\hspace{1mm}(e) ``this total world"

~

For example, a simple phrase structure grammar might define just one version of a noun phrase as in (a), using syntactic representations. But a construction grammar could also define the distinct \textsc{np}-construction in (b), further constraining the semantic domain. Thus, the utterances in (c) through (e) are noun phrases that belong to this more constrained \textsc{np}-based construction (where the semantic constraint is represented as \textsc{sem=335}).

The grammar induction algorithm used here employs an association-based beam search to identify the best sequences of slot-constraints \cite{Dunn2019}. While a grammar formalism like dependency grammar \cite{Nivre2008, Zhang2012} must identify the head and attachment type for each word, a construction grammar must identify the representation type for each slot-constraint. This leads to a larger number of potential representations and the beam search has been used to explore this space efficiently. Previous work has used the Minimum Description Length (MDL) paradigm \cite{Goldsmith2001, Goldsmith2006} to describe the fit between a grammar and a corpus as an optimization function during training.

With the exception of the use of semantic representations for slot-constraints, the meaning of constructions is not taken into account here. This is a necessary simplification. Nonetheless, it is important to remember that -- to the extent that these patterns are strong manifestations of association across slots -- it is likely that they each possess a distinct meaning as well as a distinct form. 

The experiments in this paper are centered on sub-sets of corpora containing 100k words. This is significantly less data than previous work \citep{d18}. The idea is to measure the degree to which the grammar itself changes when the induction algorithm is exposed to a more realistic amount of linguistic usage. Because the impact of training size is not clear on the MDL metric, the grammars in this paper are based on the beam search together with an MDL-based metric for choosing the optimum threshold for the $\Delta P$ association measure \citep{DunnIJCL} used in the beam search. But a final MDL-based selection stage is not employed.

Previous work represented semantic domains using word embeddings clustered into discrete categories. To provide better representations for less common vocabulary items, the embeddings here are derived from fastText \cite{Grave2019}, using k-means (the number of clusters is set to 1 per 1,000 words). Thus, the assumption is that each lexical item belongs to a single domain. Drawing on the universal part-of-speech tag-set \cite{pdm12, nn}, semantic domains are only applied to open-class lexical items, on the assumption that more functional words do not carry domain-specific information. The codebase for grammar induction is open source.\footnote{\href{https://github.com/jonathandunn/c2xg/tree/v0.03}{https://github.com/jonathandunn/c2xg}}

\section{Data and Experimental Design}

\begin{table}
\centering
\begin{tabular}{|c|l|c|}
\hline
\textbf{ID} & \textbf{Data Source} & \textbf{Condition} \\
\hline
\textsc{ac-ind} & Academic Articles & Production \\
\textsc{pg-ind} & Published Books & Production \\
\hline
\textsc{ac-agg} & Academic Papers & Perception \\
\textsc{pg-agg} & Published Books & Perception \\
\hline
\textsc{tw-agg} & Tweets & Background \\
\textsc{cc-agg} & Web Crawled & Background \\
\textsc{wi-agg} & Wikipedia Articles & Background \\
\textsc{nw-agg} & News Articles & Background \\
\hline
  \end{tabular}
  \caption{Sources of Language Data}
  \label{tab:1}
\end{table}

The basic experimental framework in this paper is to apply grammar induction to independent sub-sets of corpora drawn from different registers. We find the \textit{growth curve} of grammars and lexicons by measuring the increase in representations as these individual subsets are combined. In this case, we examine the representations learned from between 100k and 2 million words in increments of 100k, for a total of 20 observations per condition. Further, we measure the \textit{convergence} of grammars by quantifying pairwise similarities within each condition. In this framework, a \textit{condition} is defined by the data used for learning representations. For example, we examine the convergence of grammars learned from news articles by measuring pairwise similarity across 200 randomly selected combinations of unique sub-sets of the corpus of news articles.

Because of variation in registers, or varieties associated with the context of production \cite{Biber2009}, some grammatical constructions are incredibly rare in one type of corpus but quite common in another type \cite{FodorCrowther+2002+105+145, Sampson2002}. Along these same lines, some registers have more technical terms and thus a larger lexicon with more rare words. Both of these factors mean that the relationship between grammar and the lexicon could be an artifact of one particular register. To control for this possibility, the experiments in this paper are replicated across six registers, as shown in Table 1. 

First, corpora representing unique individuals are taken from academic articles and from Project Gutenberg. In this condition, each additional increment of data represents a new speaker (e.g. Dickens, followed by Austen, followed by James). Second, corpora representing aggregations of individuals are taken from the same registers; the difference here is that each additional increment of data does not represent a unique new speaker, only an increased amount of language use. Third, background corpora representing other aggregations of individuals are taken from tweets, web pages, Wikipedia articles, and news articles. These background corpora provide a baseline against which we compare variation in production-based vs perception-based models. Does any observed difference between the \textit{production} and \textit{perception} conditions fall within the expected range observed within this baseline?

In the first condition, \textit{production}, each increment of data (100k words) represents the production of a single individual. In other words, a model trained on this sub-set of the corpus is a representation of only that one individual's production. A corpus of academic articles is drawn from the field of history \cite{Daltrey2020}. This corpus represents the \textsc{ac-ind} condition, meaning the \textit{Academic} register representing \textit{Individuals}. A corpus of books from Project Gutenberg is drawn from 20th century authors. This corpus represents the \textsc{pg-ind} condition, meaning the \textit{Project Gutenberg} data organized by \textit{Individuals}. Each grammar and lexicon in this condition is trained on the production of a single speaker.

In the second condition, \textit{perception}, these production-based corpora are contrasted with data from the same registers in which each increment of 100k words represents many unknown individuals aggregated together. In other words, a model trained on this sub-set of the corpus reflects the perception of a single individual exposed to many other speakers. The academic register is represented by the British Academic Written English Corpus \cite{Alsop2009}, drawn from proficient student writing. This provides the \textsc{ac-agg} condition, representing the \textit{Academic} register but with each increment an \textit{Aggregation} of many unknown individuals. The register of books is drawn from the same Project Gutenberg corpus, this time with at most 500 words in each increment representing a single author. This ensures that there is little individual-specific information present in the corpus. This variant provides the \textsc{pg-agg} condition, representing \textit{Project Gutenberg} data as an \textit{Aggregation} of many individuals.

To provide a baseline, these paired corpora are contrasted with four further sources which represent an aggregation of many unknown individuals: social media data from tweets (\textsc{tw-agg}), web data from the Common Crawl (\textsc{cc-agg}), Wikipedia articles (\textsc{wi-agg}), and news articles, with no more than 10 articles from the same publication per increment (\textsc{nw-agg}). This range of sources ensures that the experiments do not depend on the idiosyncratic properties of a single register.

Each \textit{ID} in Table 1 represents 2 million words, divided into increments of 100k words. Representations are learned independently on each increment in isolation. In other words, the grammar induction algorithm is applied to each increment of 100k words, with no influence from the other sections of the overall corpus. Thus, each grammar simulates the representations learned from exposure to a fixed amount of language data. The \textit{amount} of exposure is held constant (at 100k words per grammar), allowing us to measure the influence of individuals (production) vs. aggregations of individuals (perception). 

The growth of grammars and lexicons is simulated by creating the union of these independent sub-sets: for example, the grammar from Dickens plus the grammar from Austen plus the grammar from James. This means that, after observing 2 million words, the production-based condition has observed the union of 20 different individuals. This design is required to represent the production-based condition because of the difficulty of finding 2 million words for many different individuals. This means that the perception-based condition at 2 million words samples from potentially tens of thousands of speakers while the production-based condition samples from just 20 speakers.

Thus, the growth curves potentially depend on the order in which different samples are observed. In other words, there is a chance that differences between growth curves are artifacts of particular orders of observation and not actual differences between corpora. To test this possibility, we simulate growth curves from 100 random samples for each condition. For each sample, we calculate the coefficient of the regression between the amount of the data and the number of representations, a measure of the growth curve. This provides a population of growth curves for each condition. We then use a t-test to determine whether this sample of growth curves represents a single population. In every case, there is no difference. This gives us confidence that the order of observations has no influence on the final results; the curves reported here are averaged across these 100 samples.

\section{Measuring Growth Curves and Grammatical Overlap}

\begin{figure*}[t]
    \centering
    \includegraphics[width=475 pt]{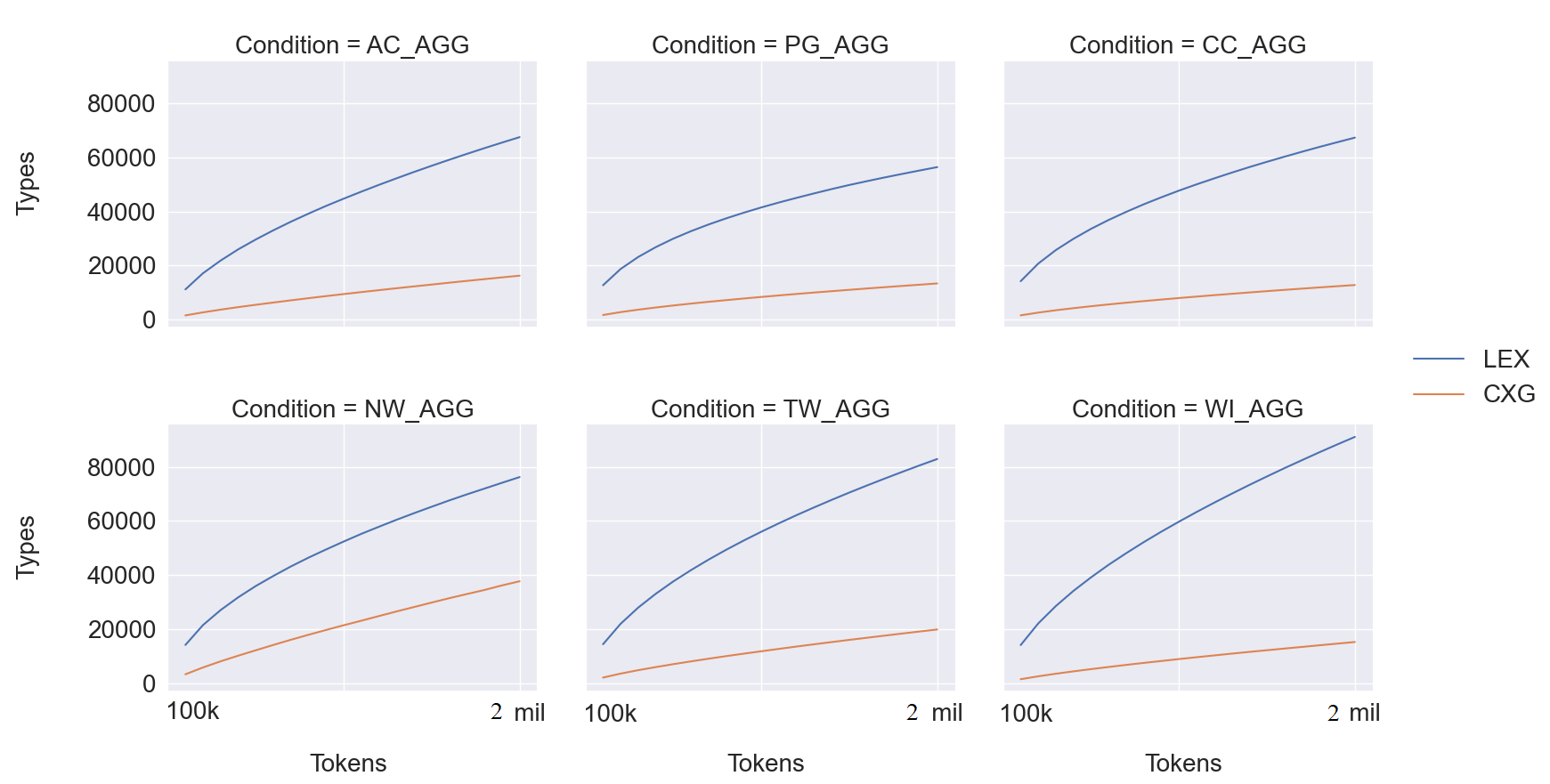}
    \caption{Growth Curve of the Lexicon Contrasted with the Grammar}
    \label{fig:lex_size}
\end{figure*}

\begin{table*}
\centering
\begin{tabular}{|ccccc|ccccc|}
\hline
\textbf{Lexicon} & ~ & ~ & ~ & ~ & \textbf{Grammar} & ~ & ~ & ~ & ~ \\
\textit{Condition} & \textbf{$\alpha$}	& \textbf{[0.025} & \textbf{0.975]} & \textbf{Max $N$} & \textit{Condition} & \textbf{$\alpha$}	& \textbf{[0.025} & \textbf{0.975]} & \textbf{Max $N$}\\
\hline
\textsc{ac-agg} & 0.776 & [0.772 & 0.782] & 67.4k & \textsc{ac-agg} & 0.660 & [0.657 & 0.664] & 16.2k \\
\textsc{pg-agg} & 0.771 & [0.764 & 0.780] & 56.3k & \textsc{pg-agg} & 0.652 & [0.652 & 0.654] & 13.3k \\
\textsc{cc-agg} & 0.782 & [0.775 & 0.790] & 67.2k & \textsc{cc-agg} & 0.649 & [0.648 & 0.651] & 12.7k \\
\textsc{nw-agg} & 0.788 & [0.782 & 0.795] & 76.2k & \textsc{nw-agg} & \textbf{0.721} & [\textbf{0.718} & \textbf{0.724}] & \textbf{37.7k} \\
\textsc{tw-agg} & 0.793 & [0.787 & 0.799] & 82.9k & \textsc{tw-agg} & 0.678 & [0.676 & 0.680] & 19.8k \\
\textsc{wi-agg} & 0.797 & [0.793 & 0.803] & 91.1k & \textsc{wi-agg} & 0.657 & [0.654 & 0.660] & 15.2k \\
\hline
  \end{tabular}
  \caption{$\alpha$ Parameters and Confidence Intervals for Growth Curve Estimation by Register}
  \label{tab:2}
\end{table*}

The growth of the lexicon is expected to take a power law distribution in which the number of lexical items is proportional to the total number of words in the corpus, as shown in (1). The challenge in understanding the rate of growth, then, is to estimate the parameter $\alpha$. The simplest method is to undertake a least-squares regression using the log of the size of the corpus and number of representations, as show in (2). On some data sets, this method is potentially problematic because fluctuations in the most infrequent representations can lead to a poor fit at certain portions of the curve \cite{Clauset2009}. We validated the experiments in this paper by conducting comparisons between estimated $\alpha$ parameters and synthesized data following Heap's law. These comparisons confirm that the traditional least-squares regression method provides an accurate measure of the growth curve.

\begin{equation}
p(x) \propto x^{-\alpha}
\end{equation}

\begin{equation}
\log p(x) = \alpha \log x + c
\end{equation}

The first question is the degree to which there is variation in the $\alpha$ parameter across representation type (grammar vs lexicon) or condition (production vs perception). For each case, such as perception-based grammar induction from news articles, we calculate the growth curve as described above using least-squares regression on the mean growth curve. We then report both the estimated $\alpha$ and the confidence interval for determining whether differences in the parameter values are significant.

\begin{equation}
d_{J}(A,B)= 1 - \frac{\left | A \cap B \right |}{\left | A \cup  B \right |}
\end{equation}

The second question is the degree to which the representations from individual sub-sets of a corpus agree with one another. To measure this, we use the Jaccard distance between grammars, shown in (3). To calculate the Jaccard distance, we first form the union of the two grammars being compared and, second, create a vector for each with binary values indicating whether a particular item is present or not present. The Jaccard distance then measures the difference between these binary vectors, with higher values indicating that there is more distance between grammars and lower values indicating that the grammars are more similar.

\section{Experiment 1. Growth Curves Across Grammar and the Lexicon}

\begin{figure*}[t]
    \centering
    \includegraphics[width=475 pt]{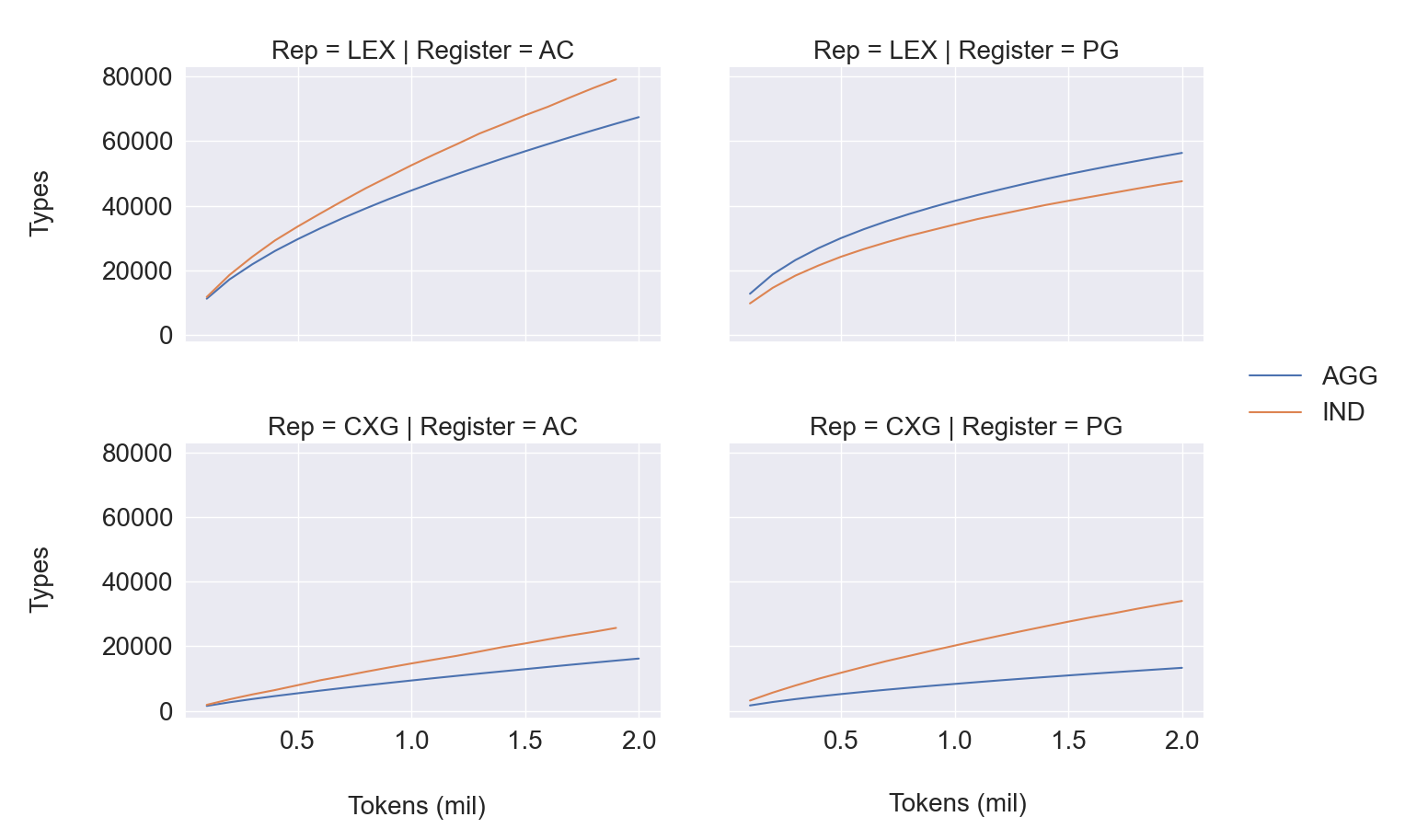}
    \caption{Growth Curves for the Production and Perception Conditions}
    \label{fig:cxg_size}
\end{figure*}

\begin{table*}
\centering
\begin{tabular}{|ccccc|ccccc|}
\hline
\textbf{Lexicon} & ~ & ~ & ~ & ~ & \textbf{Grammar} & ~ & ~ & ~ & ~ \\
\textit{Condition} & \textbf{$\alpha$}	& \textbf{[0.025} & \textbf{0.975]} & \textbf{Max $N$} & \textit{Condition} & \textbf{$\alpha$}	& \textbf{[0.025} & \textbf{0.975]} & \textbf{Max $N$}\\
\hline
\textsc{ac-agg} & 0.776 & [0.772 & 0.782] & 67.4k & \textsc{ac-agg} & 0.660 & [0.657 & 0.664] & 16.2k \\
\textsc{ac-ind} & 0.788 & [0.784 & 0.792] & 79.1k & \textsc{ac-ind} & \textbf{0.691} & [0.686 & 0.697] & \textbf{25.7k} \\
\hline
\textsc{pg-agg} & 0.771 & [0.764 & 0.780] & 56.3k & \textsc{pg-agg} & 0.652 & [0.652 & 0.654] & 13.3k \\
\textsc{pg-ind} & 0.757 & [0.751 & 0.764] & 47.5k & \textsc{pg-ind} & \textbf{0.716} & [0.714 & 0.719] & \textbf{34.0k} \\
\hline
  \end{tabular}
  \caption{$\alpha$ Parameters and Confidence Intervals for Growth Curve Estimation by Condition}
  \label{tab:3}
\end{table*}

We begin by measuring the difference between growth curves for the lexicon and for grammars. Here we compare each of the six perception-based conditions, to see the range of behaviours across registers. This is shown in Figure 1, with the x axis showing the increasing amount of data (from 100k words to 2 million words) and the y axis showing the increasing number of representations (to a max of 80k lexical items). The red line represents the grammar and the blue line represents the lexicon. Each of the perception-based conditions (i.e., each register) is represented by a separate plot.

This figure shows that the lexicon grows much more quickly than the grammar. This is somewhat expected because, even though both of them are symbols in the Construction Grammar paradigm, they are symbols of different complexity and may have different behaviors. The other important observation is that lexical items can only be terminal units in the slots of grammatical constructions, which again suggests that the number of different terminal units should be larger than the number of grammatical constructions.  

The growth of both lexicon and grammar is visualized by the slope of the lines, with a steeper curve showing quicker growth. Further, the grammar generally levels off, with the rate of growth slowing more quickly as the amount of data increases. In other words, as we observe new data, we are less likely to continuously encounter new constructions as we are to encounter new lexical items. There is general agreement across registers, except that the corpus of news articles shows a grammar that grows much more quickly, reaching a total of 37k constructions. This is a significantly larger grammar than any of the other registers. We also see variation in the lexicon, with the vocabulary on Wikipedia growing at the quickest rate.

Which of these differences are significant? We examine this in Table 2 by looking at the coefficient of a least-squares linear regression to estimate the $\alpha$ parameter, as discussed above. Each $\alpha$ is also shown with its confidence interval, outside of which the difference is taken to be significant. These regression results formalize what is visually clear from the figure: the difference between grammar and lexicon is quite significant. Because the $r^2$ values of the regression are so high \cite{Clauset2009}, it is also the case that there is a significant but less meaningful difference across registers in both types of representation. The clearest of these register-specific outliers are Wikipedia (for the lexicon) and news articles (for the grammar); only the second of these is significantly different from all other registers.

\section{Experiment 2. Perception vs Production in Growth Curves}

Our next experiment takes a closer look at the difference in the growth curves under our two conditions, production (structured around individuals) and perception (structured around aggregations of individuals). The results are shown in Figure 2, again with the growth in number of representations (types) on the y axis and the amount of data observed (tokens) on the x axis. The top row presents the lexicon and the bottom row the grammar. Finally, the blue line represents the perception condition while the red line represents the production or individual condition.

\begin{figure*}[t]
    \centering
    \includegraphics[width=475 pt]{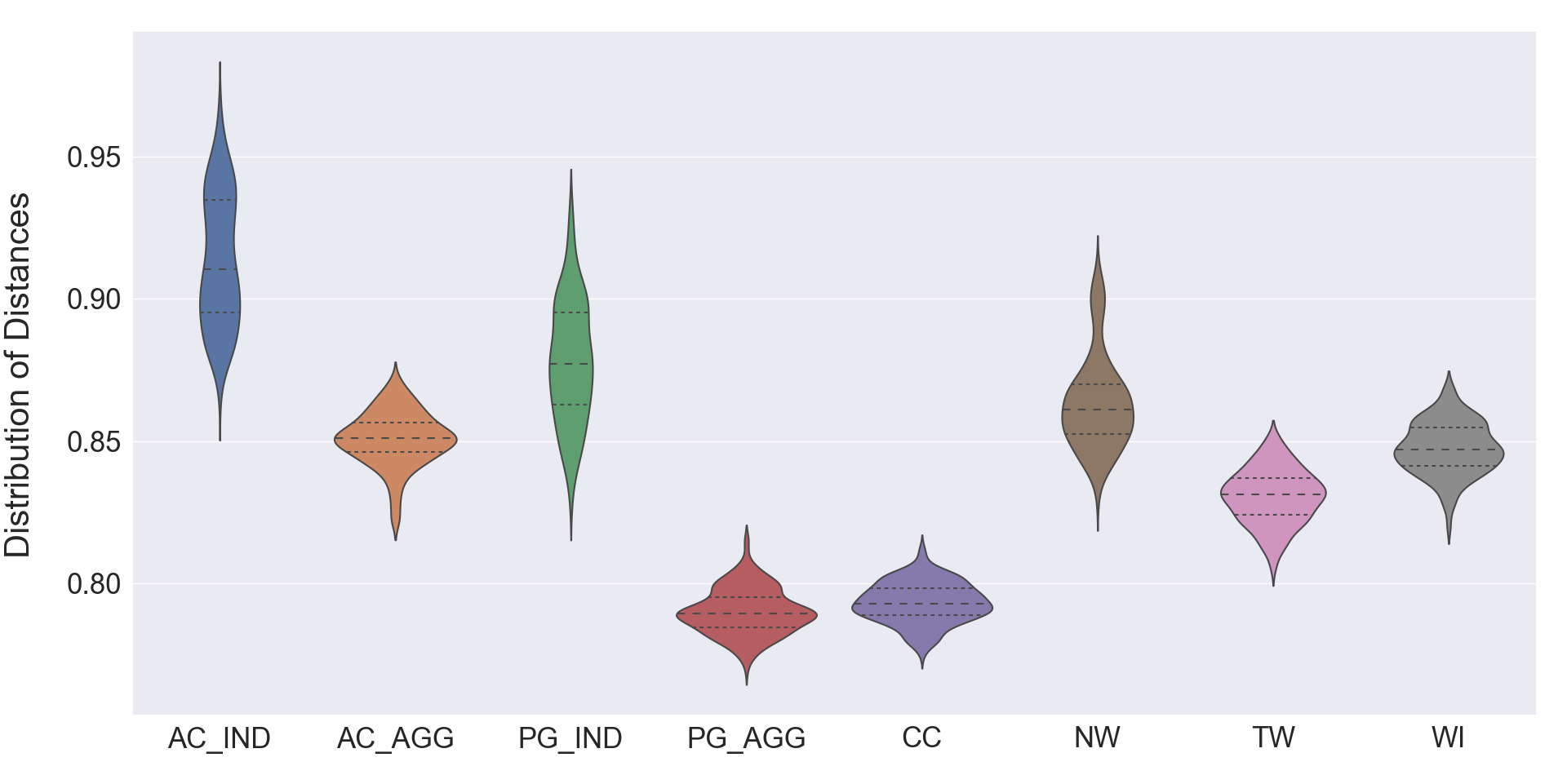}
    \caption{Distribution of Grammar Differences using Jaccard Distance}
    \label{fig:cxg_size}
\end{figure*}

The growth of the lexicon does not show any striking differences. In the academic register (\textsc{ac}), the perception condition shows a faster growth rate; but in the book register (\textsc{pg}) the reverse is true. But the growth of the grammar shows a marked difference: the production-based grammar (in red) grows more quickly in both conditions. 

This is formalized in Table 3, showing the estimated $\alpha$ parameters together with their confidence intervals for testing significance. The lexical differences, confirming what we see visually, are not significantly different in either register (i.e., the confidence intervals overlap, or very nearly do). So the difference between production and perception has no influence on the growth of the lexicon.

And yet the growth of the grammar across these two conditions is significantly different in both registers, with an especially large difference in the register of published books (\textsc{pg}). This significance is shown by the confidence intervals on the estimation of the $\alpha$ parameter; but it is also shown in the final size of the grammars: 16.2 and 13.3k (\textsc{agg}) vs 25.7k and 34.0k (\textsc{ind}). In other words, given access to data from just one individual, the grammar contains more constructions than an equal amount of data from an aggregation of individuals.

It is important to remember that the grammar induction algorithm is applied independently to each sub-set of the data. What this result shows, then, is that there are considerable individual differences or idiosyncrasies in the grammar but not in the lexicon. In both registers, grammar induction based on the production of individuals acquires more constructions given the same amount of exposure. This is important because most computational approaches to language learning assume that speakers generalize toward a single shared grammar. This implies, incorrectly, that the presence of many speakers in the training corpora is irrelevant, perhaps with the further constraint that each training corpus should represent a single community and register (like written British English). 

\section{Experiment 3. Perception vs Production in Grammar Similarity}

The previous experiments have focused on the \textit{size} and growth of the grammars without focusing on the presence of individual representations (i.e., constructions). To what degree do the grammars from each sub-set of a corpus overlap? Is there a significant difference between the overlap of perception-based and production-based representations? The basic idea in this experiment is to take a closer look at the higher growth curve in production-based grammars identified in the previous experiment: it is possible that a few of the grammars are unique, thus contributing to a higher growth curve, without a pervasive uniqueness distributed across all of the production-based grammars. 

This experiment consists in creating pairs of grammars under the two conditions. First, we sample 200 pairs drawn from each condition/register: for example, a pair from different sub-sets of the corpus of news articles. Second, we use Jaccard distance to measure the similarity of each pair. Each comparison is made within a single register, thus controlling for the possibility of register variation. This provides a broader population of pairwise similarities, allowing us to measure the uniqueness of individual grammars in each condition.

We visualize the distribution of grammar similarities using a violin plot in Figure 3. The distance measure ranges from 1 (no overlap) to 0 (complete overlap). The violin plot here shows the distributions, with width representing the density for a particular value and height representing the range of values. This shows, for example, that the \textsc{ac-ind} condition is not normally distributed. Rather, it has a large range of values with two slight peaks. The \textsc{ac-agg} condition, however, is normally distributed, with a large peak at its mean (shown here by the dotted line in the center).

The values for the Jaccard distances show that, independently of condition, these pairs of grammars are relatively dissimilar. There are many reasons why this is the case, ranging from the amount of data used to train each grammar to the possibility that constructional representations overlap with slightly different slot-constraints. Putting aside the baseline similarity that is observed using this particular measure, the larger point is that there is a clear distinction between production-based and perception-based grammars.

This figure shows a clear distinction between the production-based (\textsc{ind}) and perception-based (\textsc{agg}) conditions. The grammars learned from individuals vary widely among themselves: some pairs have a high overlap but others a low overlap. Furthermore, the most similar pairs in the individual conditions are as similar or less similar than the average pair for the aggregated condition. This indicates that there are individual differences in these grammars, the same phenomenon that resulted in the higher growth curves identified in the second experiment above.

The perception-based grammars, however, have a low degree of variation: the similarity measures are centered densely around the mean because most grammars have the same degree of similarity. This means that the aggregated or perception-based condition is forcing the induction algorithm to converge onto more stable representations by exposing it to many individuals. The inverse of this generalization is that individuals have unique or idiosyncratic constructions which are only revealed when the training corpus is centered around that individual. This finding fits well with studies in variation \cite{Dunn2019, Dunn2019a} which reveal the high degree of syntactic differences across speech communities.

We also notice in Figure 3 that the news register, although part of the perception-based condition, is not as densely centered as the other background registers. This shows the importance of including many registers in a study like this. The likely reason is that different publications enforce their own stylistic conventions. This data set is balanced to ensure that no single publication venue accounts for more than 10 of the articles in any sub-set of the corpus. It remains the case, however, that the presence of a publication-specific style may simulate a different distribution of grammar overlap.

We formalize this violin plot in Table 4 using Bayesian estimates of the mean and variance for each condition at a 99\% confidence interval. Because the Jaccard distance is between 0 and 1, we multiply each value by 100 to make the values easier to read. First, the mean distance in the production-based condition is significantly higher in each case; further, the production-based conditions have a higher mean than any of the background conditions. Second and more importantly, the variance for the production-based conditions is greater by an order of magnitude than all other conditions. Only the news register is close; and this is still more similar to the other background data sets than to the individual data sets. The variance is important because it represents the range of overlap caused by individual differences in the grammars.

These Bayesian estimates reinforce the visualization and show that there is more variance and thus more individual differences within grammars that are trained from the production of a single individual. This experiment thus confirms what is suggested by the increased growth curves seen in the second experiment: production-based grammars diverge into more individual-specific representations.

\begin{table}
\centering
\begin{tabular}{|ccc|}
\hline
\textbf{Condition} & \textbf{Mean} & \textbf{Variance}\\
\hline
\textsc{ac-ind} & 91.35 & 0.053 \\
\textsc{pg-ind} & 87.79 & 0.045 \\
\hline
\textsc{ac-agg} & 85.08 & 0.009 \\
\textsc{pg-agg} & 79.01 & 0.006 \\
\hline
\textsc{cc-agg} & 79.33 & 0.005 \\
\textsc{tw-agg} & 83.06 & 0.009 \\
\textsc{wi-agg} & 84.76 & 0.008 \\
\textsc{nw-agg} & 86.33 & 0.026 \\
\hline
  \end{tabular}
  \caption{Estimated Mean and Variation at Bayesian Confidence Interval of 99\% (Each *100 for readability)}
  \label{tab:4}
\end{table}

\section{Discussion and Conclusions}

The three computational experiments in this paper have shown that there is a significant difference between perception-based and production-based grammar induction, even when these conditions are contrasted across many registers. Grammars based on individuals (i) have a significantly steeper growth curve and (ii) a significantly more long-tailed distribution of pairwise similarity. We have also seen that the growth curve of the grammar in general does not have the same $\alpha$ parameter as the lexicon, but does still conform to the generalizations provided by Heap's Law. This supports the idea of a continuum  between grammar and the lexicon, with the symbolic representations in the grammar more complex and more abstract, thus showing a slower growth curve.

The results obtained by the three experiments overall reveal that, given a certain number of word tokens, the number of constructions extracted is higher if the sample is taken from one unique individual as opposed to a set of unknown individuals. For example, 100k words of data from academic prose written by the same individual contain 1845 construction types, while the same amount of data from a combination of  individuals contains about 1512 construction types, a difference of 333. This is not a trivial result: as a counter-factual, it would also be plausible to expect that the aggregated data would contain a wider variety of constructions because it represents a wider variety of individuals. These results therefore suggest that the constructions that are normally observed in traditional (aggregated) corpora are just the tip of the iceberg: there are many individual-specific constructions that are never observed in aggregated production. In other words, the \textit{uniqueness} of individual construction grammars is disguised when observing the aggregated usage of many individuals.

These findings are consistent with the usage-based proposal that the general grammatical representation of a language emerges as a complex-adaptive system \cite{Beckner2009}. The grammars learned in the perception-based condition contain fewer construction types and are relatively similar to each other. However, these seemingly homogeneous grammars are in fact formed from the shared usage across a number of different individuals. And, as shown in the production-based condition, these aggregated individuals on their own are likely to use very different grammars.


\bibliography{naacl2021}
\bibliographystyle{acl_natbib}

\end{document}